\documentclass[runningheads, hidelinks]{llncs}
\usepackage{lineno}

\usepackage{graphicx}
\usepackage{relsize}
\usepackage{booktabs} 
\usepackage{multirow}
\usepackage{pdfpages}
\usepackage{amsmath}
\usepackage{algorithm}
\usepackage{algorithmic}
\usepackage{todonotes}
\usepackage{enumitem}
\usepackage{ragged2e,array,url,graphicx,csquotes,hyperref,adjustbox,longtable}
\usepackage[english]{babel}
\graphicspath{ {pictures/} }

\usepackage{setspace}

\definecolor{PrologPredicate}{RGB}{0,0,200}
\definecolor{PrologVar}      {RGB}{145,032,039}
\definecolor{PrologComment}  {RGB}{169,082,044}
\definecolor{PrologOther}    {rgb}{0.2,0.2,0.2}
\definecolor{PrologString}   {rgb}{0.2,0.2,0.2}

\usepackage{listings} 
\usepackage{textcomp}

\lstnewenvironment{scasp}[1][]{\lstset{style=MySCASP,#1}}{}
\lstdefinestyle{MyInline}
{
  basicstyle = \ttfamily\color{PrologVar},
  breaklines = true,
  breakatwhitespace=true,
  upquote = true,
  literate =
  {,}{}{0\discretionary{,}{}{,}}
  {\ │}{{$\mid$}}1 
  {|}{{$\mid$}}1
  {\\\{}{{\{}}1
  {\\\}}{{\}}}1
  {[}{{\small[}}1
  {]}{{\small]}}1
  {.=.}{{\#=}}3
  {.<.}{{\#<}}3
  {.>.}{{\#>}}3
  {.=<.}{{\#=<}}4
  {.>=.}{{\#>=}}4
  {\\=}{{\char"5C=}}2
  {?-}{{?-}}2
  {:-}{{:-}}2
  {\\$}{{\$}}1
}
\lstdefinestyle{tree}
{
  basicstyle = \scriptsize\ttfamily\color{PrologPredicate},
  basewidth = 0.46em,
  moredelim = {[s][\color{PrologString}]{ \{}{\} }},
  moredelim = {*[s][{\color{PrologVar}}]{(}{)}},
  literate     =
  {.\\=.}{{\ \char"5C=\ }}3
  {\\=}{{\ \char"5C=\ }}3
  {.<.}{{\ \#<\ }}4
  {.>.}{{\ \#>\ }}4
  {.=.}{{\ \#=\ }}3
  {.=<.}{{\ \#=<\ }}5
  {.>=.}{{\ \#>=\ }}5
}
\lstdefinestyle{Plain}
{
  keywords = {},
  upquote = true,
  basicstyle = \relsize{-0.5}\ttfamily\color{PrologPredicate},
  basewidth = 0.48em,
  numbers=none,
  xleftmargin=0cm,
  moredelim = {*[s][\color{black!40!PrologPredicate}]{\#pred}{.}},
  moredelim = {*[s][\color{black!40!PrologPredicate}]{\#show}{.}},
  moredelim = {*[s][\color{black!40!PrologPredicate}]{\#hide}{.}},
  moredelim = {*[s][\color{PrologVar}]{(}{)}},
  moredelim = {*[s][\color{PrologString}]{'}{'}},
 commentstyle = \mdseries\color{PrologComment},
  morecomment=[l]\%,
   literate     =
  {|}{{$\mid$}}1
  {\\$}{{\$}}1
  {\ │}{{$\mid$}}1,
}
\lstdefinestyle{MySCASP}
{
  keywords = {},
  basicstyle = \relsize{-0.5}\ttfamily\color{PrologPredicate},
  basewidth = 0.48em,
  moredelim = {*[s][\color{PrologOther}]{:-}{.}},
  moredelim = {*[s][\color{PrologVar}]{(}{)}},
  commentstyle = \mdseries\color{PrologComment},
  morecomment=[l]\%,
  literate     =
  {&(}{{\color{PrologOther}(}}1
  {&)}{{\color{PrologOther})}}1
}
\newcommand{\blist}{\smallskip\begin{list}{$\bullet$}{\topsep=1pt \parsep=0pt \itemsep=4pt}}
\newcommand{\elist}{\end{list}\medskip}

\newcommand{\bnum}{\begin{list}{}{\topsep=2pt \parsep=0pt \itemsep=1pt}}
\newcommand{\enum}{\end{list}\medskip}

\begin{document}

\title{Using Logic Programming and Kernel-Grouping for Improving Interpretability of Convolutional Neural Networks}
\author{%
  Parth Padalkar$^1$\orcidID{0000-0003-1015-0777}, \quad
  Gopal Gupta$^1$\orcidID{0000-0001-9727-0362} \\
  \institute{$^1$The University of Texas at Dallas, Richardson, USA \\ 
  \email{$^1$\{parth.padalkar, gupta\}@utdallas.edu}
  }
}
\authorrunning{Parth Padalkar and Gopal Gupta}

\maketitle              
%


\begin{abstract}
Within the realm of deep learning, the interpretability of Convolutional Neural Networks (CNNs), particularly in the context of image classification tasks, remains a formidable challenge. To this end we present a neurosymbolic framework, NeSyFOLD-G that generates a symbolic rule-set using the last layer kernels of the CNN to make its underlying knowledge interpretable.
What makes NeSyFOLD-G different from other similar frameworks is that we first find groups of similar kernels in the CNN (kernel-grouping) using the cosine-similarity between the feature maps generated by various kernels. Once such kernel groups are found, we binarize each kernel group's output in the CNN and use it to generate a binarization table which serves as input data to FOLD-SE-M which is a Rule Based Machine Learning (RBML) algorithm. FOLD-SE-M then generates a rule-set that can be used to make predictions. We present a novel kernel grouping algorithm and show that grouping similar kernels leads to a significant reduction in the size of the rule-set generated by FOLD-SE-M, consequently, improving the interpretability. 
This rule-set symbolically encapsulates the connectionist knowledge of the trained CNN. The rule-set can be viewed as a \textit{normal logic program} wherein each predicate's truth value depends on a kernel group in the CNN. Each predicate in the rule-set is mapped to a concept using a few semantic segmentation masks of the images used for training, to make it human-understandable. The last layers of the CNN can then be replaced by this rule-set to obtain the NeSy-G model which can then be used for the image classification task. The goal directed ASP system s(CASP) can be used to obtain the justification of any prediction made using the NeSy-G model. We also propose a novel algorithm for labeling each predicate in the rule-set with the semantic concept(s) that its corresponding kernel group represents.

\end{abstract}

\section{Introduction}
Interpretability of deep learning models is an important issue that has resurfaced in recent years as these models have become larger and are being applied to an increasing number of tasks. Some applications such as autonomous vehicles \cite{kanagaraj2021}, disease diagnosis \cite{Sun2016ComputerAL}, and natural disaster prevention \cite{Ko2012disaster} are very sensitive areas where a wrong prediction could be the difference between life and death. The above tasks rely heavily on good image classification models such as Convolutional Neural Networks (CNNs). A CNN is a deep learning model used for a wide range of image classification and object detection tasks, first introduced by Y. Lecun et al. \cite{cnn}. Current CNNs are extremely powerful and capable of outperforming humans in image classification tasks. A CNN is inherently a blackbox model, though attempts have been made to make it more interpretable \cite{zhang2017growing,zhang2018interpreting}. There is no way to tell whether the predictions made by the model are based on concepts meaningful to humans, or are simply the outcome of coincidental correlations. If the knowledge of the trained CNN becomes interpretable then domain experts can scrutinize this knowledge and point out any biases or spurious correlations that the CNN might have learnt which could lead to wrong predictions. Thus retraining with better and more targeted data can be suggested by the experts.

We propose a framework for interpretable image classification using CNNs called NeSyFOLD-G. A CNN, like any deep neural network is composed of multiple layers. We focus on the convolution layer, more specifically the last convolution layer of a CNN in this work. The convolution layer is composed of kernels. 

A kernel, also known as a filter, is a 2D matrix. It acts like a small, specialized magnifying glass that slides over an image to help recognize specific features or patterns in the image, like edges, curves, or textures. It does this by multiplying its values with the pixel values of the image in a small region, and then it adds up those products. This process helps highlight important parts of the image. As the kernel slides over the entire image, it creates a new, simplified version of the image that emphasizes the patterns it's looking for. This simplified version is called a feature map. The CNN then uses these feature maps to understand the image and make predictions.

The NeSyFOLD-G framework can be used to create a \textit{NeSy-G} model which is a composition of the CNN and a rule-set generated from kernels in its last convolution layer. A Rule Based Machine Learning (RBML) algorithm called FOLD-SE-M \cite{foldsem} is used for generating the rule-set by using binarized outputs of the groups of similar kernels in a trained CNN. The rule-set is a default theory represented as a normal logic program, \cite{Lloyd87} i.e., Prolog extended with negation-as-failure. The binarized output ($0/1$) of the kernel groups influences the truth value of the predicates appearing in the rule body.  The rule-set can also be viewed as a stratified Answer Set Program and the s(CASP) \cite{scasp} ASP system can be used to obtain justifications of the predictions made by the NeSy-G model. The rule-set also serves as a global explanation for the predictions made by the CNN.

Our first novel contribution is the \textit{kernel grouping algorithm} that finds groups of similar kernels in the CNN based on the cosine similarity score of their corresponding generated feature maps.

\begin{figure}
    \centering
    \includegraphics[width=1\linewidth, height = 8cm]{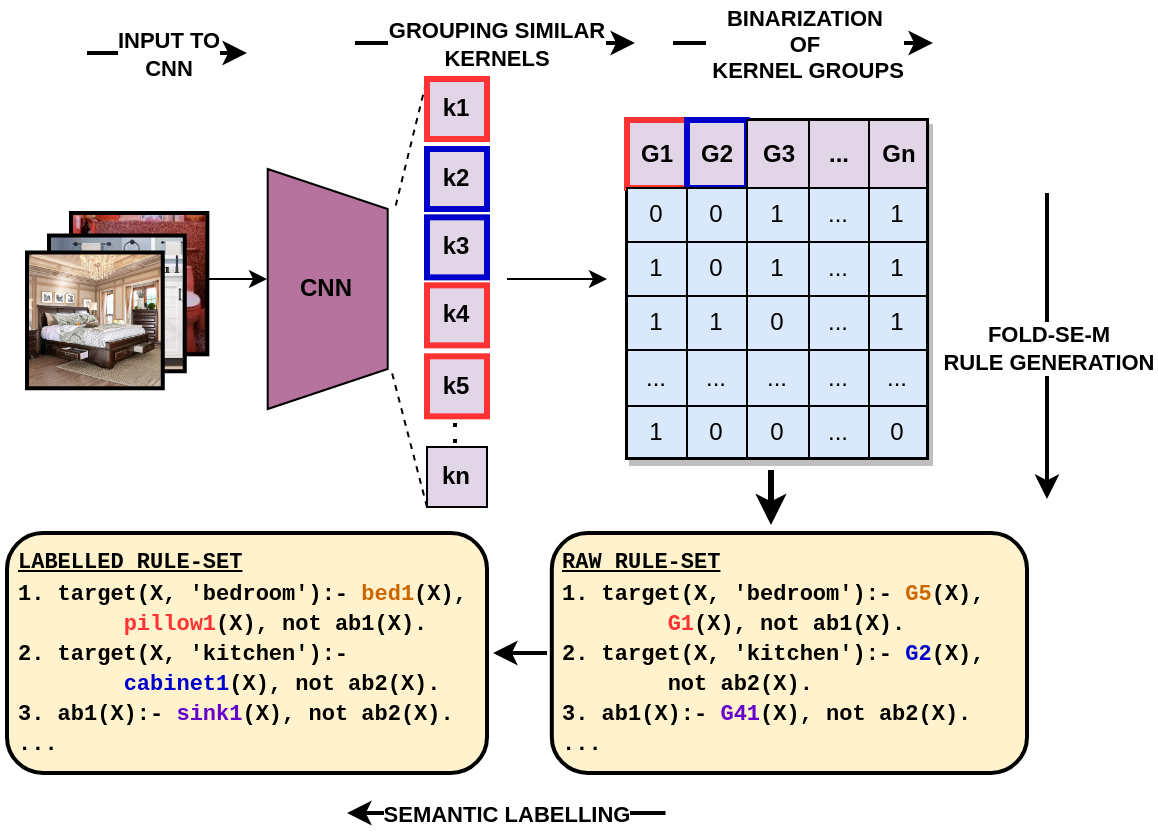}
    \caption{The NeSyFOLD-G framework. Each kernel group is depicted with a unique color in the rule-set.}
    \label{fig_1}
\end{figure}

We also introduce a semantic labelling algorithm that can be used to label the predicates in the rule-set with the semantic concept(s) that their corresponding kernel groups represent in the images. For example, the predicate {\tt 52(X)} corresponding to kernel group {\tt 52} in the last convolution layer of the CNN will be replaced by {\tt bathtub(X)} in the rule-set, if kernel group {\tt 52} has learnt to look for ``bathtubs" in the image. Fig. \ref{fig_1} illustrates the NeSyFOLD-G framework.

Padalkar et al. proposed the NeSyFOLD framework \cite{nesyfold} which shares similarities with the NeSyFOLD-G framework. The major difference that separates NeSyFOLD-G from NeSyFOLD is that the truth values of predicates in the generated rule-set is influenced by the binarized output of \textit{groups} of similar kernels. In NeSyFOLD each predicate's truth value is influenced by single kernels in the CNN. 
However, it is known that groups of kernels in the last layer are responsible for representing a single concept. Yang et al. \cite{lavise} proposed an attention-based masking mechanism for finding the concept learnt by a single kernel by accounting for the other kernels with similar attention weights. Their approach serves as motivation behind our kernel grouping algorithm that uses the cosine similarity score between feature maps of various kernels to find the groups of similar kernels.

The size of the rule-set generated can be used as a metric for interpretability. Lage et al. \cite{rulesetinterpretability} comprehensively showed through human evaluations that as the size of the rule-set increases the difficulty in interpreting the rule-set also increases. Padalkar et al. show that NeSyFOLD framework generates a smaller rule-set than the ERIC system \cite{eric} which was the previous SOTA. We show that NeSyFOLD-G, achieves a significant reduction in the size of the rule-set generated while maintaining or improving on the accuracy and fidelity in comparison to the NeSyFOLD framework.

To summarize, our contributions are as follows:

\begin{enumerate}
    \item We present a novel kernel grouping algorithm that constitutes the heart of the NeSyFOLD-G framework for improving interpretability of the generated rule-set.
    \item We also introduce a semantic labeling algorithm for labeling the predicates of the rule-set generated by the NeSyFOLD-G framework.
\end{enumerate}

\section{Background}
\medskip\noindent\textbf{FOLD-SE-M:}
The FOLD-SE-M algorithm \cite{foldsem} that we employ in our framework, learns a rule-set from data as a \textit{default theory}. 
Default logic is a non-monotonic logic used to formalize commonsense reasoning. A default $D$ is expressed as:
  
\begin{equation}\label{eq_1}
    D = A: \textbf{M} B \over\Gamma    
\end{equation}

\noindent Equation \ref{eq_1} states that the conclusion $\Gamma$ can be inferred if pre-requisite $A$ holds and $B$ is justified. $\textbf{M} B$ stands for ``it is consistent to believe $B$".
Normal logic programs can encode a default theory quite elegantly \cite{gelfondkahl}. A default of the form: 
$$\alpha_1 \land \alpha_2\land\dots\land\alpha_n: \textbf{M} \lnot \beta_1, \textbf{M} \lnot\beta_2\dots\textbf{M}\lnot\beta_m\over \gamma$$
\noindent can be formalized as the
normal logic programming rule:
$$\gamma ~\texttt{:-}~ \alpha_1, \alpha_2, \dots, \alpha_n, \texttt{not}~ \beta_1, \texttt{not}~ \beta_2, \dots, \texttt{not}~ \beta_m.$$
\noindent where $\alpha$'s and $\beta$'s are positive predicates and \texttt{not} represents negation-as-failure. We call such rules \emph{default rules}. 
Thus, the default 

$$bird(X): M \lnot penguin(X)\over flies(X)$$

\noindent will be represented as the following default rule in normal logic programming:

{\tt flies(X) :- bird(X), not penguin(X).}

\noindent We call {\tt bird(X)}, the condition that allows us to jump to the default conclusion that {\tt X} flies, the {\it default part} of the rule, and {\tt not penguin(X)} the \textit{exception part} of the rule. 

FOLD-SE-M \cite{foldsem} is a Rule Based Machine Learning (RBML) algorithm. It generates a rule-set from tabular data, comprising rules in the form described above. The complete rule-set can be viewed as a stratified answer set program. It uses special {\tt abx} predicates to represent the exception part of a rule where {\tt x} is unique numerical identifier.   
FOLD-SE-M incrementally generates literals for \textit{default rules} that cover positive examples while avoiding covering negative examples. It then swaps the positive and negative examples and calls itself recursively to learn exceptions to the default when there are still negative examples falsely covered.

There are $2$ tunable hyperparameters, $ratio$, and $tail$. 
The $ratio$ controls the upper bound on the number of false positives to the number of true positives implied by the default part of a rule. The $tail$ controls the limit of the minimum number of training examples a rule can cover.
FOLD-SE-M generates a much smaller number of rules than a decision-tree classifier and gives higher accuracy in general.

\section{Learning}
In this section we describe the process of generating a rule-set from the CNN and obtaining the NeSy-G model.
We start by training the CNN on the input images for the given image classification dataset. Any optimization technique can be used for updating the weights. Fig. \ref{fig_1} illustrates the learning pipeline.

\medskip\noindent\textbf{Binarization:}
Once the CNN has been fully trained to convergence, we pass the full training set consisting of $n$ images to the CNN. For each image $i$ in the training set, let $A_{i,k}$ denote the feature map generated by kernel $k$ in the last convolutional layer. The feature map $A_{i,k}$ is a $2D$ matrix of dimension determined by the CNN architecture. For each image $i$ there are $K$ feature maps generated where $K$ is the total number of kernels in the last convolutional layer of the CNN.
To convert each of the feature maps to a single value we take the norm of the feature maps as demonstrated by eq. \eqref{eq_2} to obtain $a_{i,k}$.

\medskip\noindent\textit{Kernel grouping algorithm: }We then find the groups of similar kernels in the CNN. Consider a kernel $\hat{k}$ for which we need to identify the most similar kernels. We do this by first finding the \textit{top-10} images $\hat{i}_1, \hat{i}_2, ..., \hat{i}_{10}$ that activate $\hat{k}$ the most, according to the norm values of the feature maps generated by $\hat{k}$ for these images. Now, we compute the cosine similarity score between $A_{\hat{i}_{g}, \hat{k}}$ and $A_{\hat{i}_{g}, k^\prime}$, where g $\in [1,10]$ and $k^\prime$ is some kernel in the last layer the last layer of the CNN. The similarity score of kernel $k^\prime$ w.r.t $\hat{k}$ is calculated by taking the mean of the cosine similarity scores for all the \textit{top-10} images $\hat{i}_1, \hat{i}_2, ..., \hat{i}_{10}$ as $sim_{\hat{k}, k^\prime}$. The similarity score is a value between $0$ and $1$. Thus, we calculate the similarity score of all kernels in the last layer of the CNN w.r.t to $\hat{k}$. The group of kernel $\hat{k}$ would then constitute of all kernels that have a similarity score w.r.t $\hat{k}$ greater than a user-defined similarity threshold $\theta_s$. 

Hence, we find a group of similar kernels $G_k$ for all the kernels $k$ in the last layer of the CNN. Note that the total number of kernel groups $G_k$ is the same as the total number of kernels in the last layer of the CNN. 

Next, for each kernel group $G_k$ we obtain the group norm $a_{i, G_k}$ for each image $i$ in the training set. This is achieved by taking the mean of the norms corresponding to each kernel in $G_k$ for each image $i$. This leads to the creation of a table $T_G$ with each row representing an image and each column representing the group norm for each of the kernel groups $G_k$.

Finally, for each kernel group we convert the group norm values to either $0$ or $1$ which symbolizes the kernel group ``activating" or ``deactivating" for each image. This is called \textit{binarization} of the kernel groups. This is done by determining an appropriate threshold $\theta_{G_k}$ for each kernel group $G_k$ to binarize its output. The threshold $\theta_{G_k}$ is calculated as a weighted sum of the mean and the standard deviation of the group norms $a_{i,G_k}$ for all images $i$ in the training set, denoted by eq. \eqref{eq_3} where $\alpha$ and $\gamma$ are user-defined hyperparameters.

Thus a binarization table $B_G$ is created. Each row in the table represents an image and each column is the binarized kernel group value represented by either a $0$ if $a_{i, G_k} \leq \theta_{G_k} $ or $1$ if $a_{i, G_k} > \theta_{G_k} $ (\textit{cf.} Fig. \ref{fig_1} (right)).

\begin{align}
    a_{i,k} =& ||A_{i,k}||_2 \label{eq_2}\\
    \theta_{G_k} = & \alpha \cdot \overline{a_{G_k}} + \gamma\sqrt{\frac{1}{n}\sum(a_{i,G_k} - \overline{a_{G_k}})^2} \label{eq_3}
\end{align}

\medskip\noindent\textbf{Rule-set Generation:} The binarization table $B_G$ is given as an input to the FOLD-SE-M algorithm to obtain a rule-set in the form of a normal logic program. The FOLD-SE-M algorithm finds the most influential features in the $B_G$ and generates a rule-set that has these features as predicates. Since $B_G$ has features as kernel group ids, the raw rule-set has predicates with names in the form of their corresponding kernel group's id.
An example rule could be:\\
\indent\indent {\tt target(X,`2') :- not 3(X), 54(X), not ab1(X).}\\
This rule can be interpreted as ``Image X belongs to class {\tt`2'} if kernel group {\tt 3} is not activated and kernel group {\tt 54} is activated and the abnormal condition (exception) {\tt ab1} does not apply". There will be another rule with the head as {\tt ab1(X)} in the rule-set. The binarized output of a kernel group would determine the truth value of its predicate in the rule-set.
The rule-set generated is in the form of a decision list, i.e., the next rule is checked only if the current rule and all the rules above it were not satisfied. 

\medskip\noindent\textbf{Semantic labeling:}  Groups of kernels activate in synergy to identify concepts in the CNN. Since we capture the outputs of the kernel groups as truth values of predicates in the rule-set, we can label the predicates with the semantic concept(s) that the corresponding kernel group has learnt. 
Thus, the same example rule from above may now look like:\\
\indent\indent{\tt target(X,`bathroom') :- not bed(X), bathtub(X), not ab1(X).}

\noindent We introduce a novel semantic labelling algorithm to automate the semantic labelling of the predicates in the rule-set generated. The details of the algorithm are discussed later.

The NeSy-G model is conceptualized as the model obtained after replacing all the layers following the last convolutional layer with the rule-set generated by applying the FOLD-SE-M algorithm on the binarization table $B_G$.

\section{Inference}
For using the NeSy-G model to obtain predictions on
the test set, we first obtain the kernel feature maps for each kernel in the last convolutional layer. Then, we compute the group norms for all kernel groups that were found in the learning process to obtain the table $T^{test}_{G_k}$. From $T^{test}_{G_k}$ we obtain the binarization table $B^{test}_{G}$ by binarizing the output of each kernel group in $T^{test}_{G_k}$ by using the threshold $\theta_{G_k}$ calculated in the learning phase. Next, for each binarized vector $b$ in $B^{test}_{G}$, we use the labeled/unlabelled rule-set obtained in the learning phase to make predictions. The truth value of the predicates in the rule-set is determined by the corresponding binarized kernel group values in $b$. FOLD-SE-M toolkit's built-in rule-interpreter can be used to obtain the predicted class of $b$ given the rule-set. The binarized kernel group values in $b$ can also be listed as facts and the rule-set which can be viewed as a stratified answer set program, can be queried with the s(CASP) interpreter \cite{scasp} to obtain the justification as well as the target class. Note that s(CASP) searches for the answer set in a goal directed manner, which implies that the rules are checked from the top to the bottom one by one. Hence, the first answer set that is found to satisfy the rule-set with the given facts entails the intended prediction made by the NeSy-G model.

\section{Semantic Labelling of Predicates}
The raw rule-set generated by FOLD-SE-M initially has kernel group ids as predicate names.  Also, since the FOLD-SE-M algorithm finds only the most influential kernel groups, the number of kernel groups that actually appear in the rule-set is usually very low in comparison to the total number of kernel groups. We present a novel algorithm for automatically labelling the corresponding predicates of the kernel groups with the semantic concept(s) that the kernel groups represent.

    

\begin{figure}[h!]
    \centering
    \includegraphics[width=1\linewidth, height = 7cm]{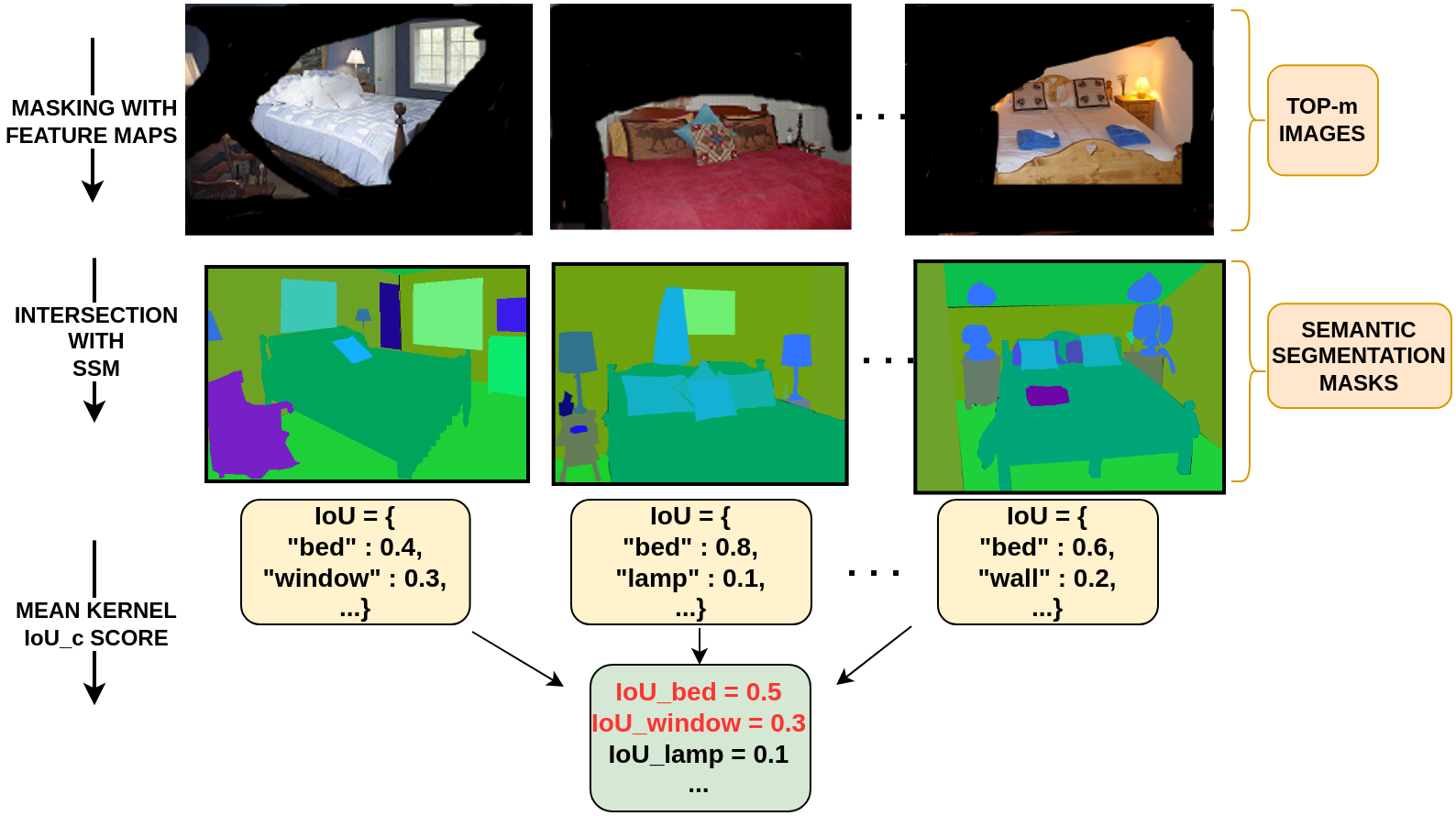}
    \caption{The calculation of mean $IoU_c$ scores for a kernel.}
    \label{fig_2}
\end{figure}

Xie et al. \cite{conceptsinCNN} showed that each kernel in the CNN may learn to represent multiple concepts in the images. Hence each kernel group may also represent multiple concepts.  As a result, we assign semantic labels to each predicate, denoting the names of the semantic concepts learnt by the corresponding kernel group. 
To regulate the extent of approximation, i.e., to dictate the number of concept names to be included in the predicate label, we introduce a hyperparameter 
\textit{margin}.
This hyperparameter exercises control over the precision of the approximation achieved. Figure \ref{fig_2} illustrates the semantic labelling of a given predicate. The algorithm requires a dataset that has semantic segmentation masks of the training images. This essentially means that for every image $i$ in the dataset $I$, there is an image $i_M$ where every pixel is annotated with the label of the object (concept) that it belongs to (Fig. \ref{fig_2} middle). We denote these by $I_M$.

The CNN that is trained on the training set is used to obtain the norms $a_{i,k}$ of the feature maps $A_{i,k}$ generated by each kernel $k$ in the last convolution layer. Next, as the respective kernel groups for each kernel are known, the table $T^{I_m}_{G_k}$ is created where each row represents the images whose corresponding semantic segmentation masks are available and the columns are the kernel group norms.

Now, consider some kernel group $G_{\hat{k}}$ that has $l$ kernels in the group namely, $\hat{k}_1, \hat{k}_2,...,\hat{k}_l$ . The \textit{top-m} images $i^{\prime}_1, i^{\prime}_2,...,i^{\prime}_m \in I^{\prime}_m$, according to the group norm values are selected. We need to calculate the group's 
\textit{Intersection over Union ($IoU_c$)} score for each concept $c$ visible in the \textit{top-m} images that most activate the group. Then according to this score for each concept $c$, the label of the kernel group's predicate should comprise of the top concepts that the kernel group is detecting.
\begin{align}
     IoU_c(i^{Mask}, i) =& \frac{\text{no. of non-zero pixels in } c \cap i}{\text{no. of non-zero pixels in } i} \label{eq_5}
 \end{align}

For a given image  $i_j \in I^{\prime}_m$, the resized feature map generated by every kernel in the kernel group is used to mask the image to obtain $i^{\hat{k}_1}_j, i^{\hat{k}_2}_j,...,i^{\hat{k}_l}_j$. Fig. \ref{fig_2} (top) shows a few images masked with the resized feature maps generated by a kernel. For each of these masked images, the $IoU_c$ score is calculated using eq. \eqref{eq_5} for each concept $c$, that appears in the corresponding semantic segmentation mask $i^{Mask}_j$ of the image $i_j$. Fig. \ref{fig_2} (middle) shows the semantic segmentation masks of the images at the top. Next, each kernel's $IoU_c$ score for all the \textit{top-m} images, for all concepts $c$ is calculated. Each kernel's mean 
$IoU_c$ score is calculated by taking the mean score over all images. Finally, the kernel group's $IoU_c$ score is calculated by taking the mean of the mean $IoU_c$ score of each kernel for each concept $c$.

The algorithm can be summarized as follows:
\begin{enumerate}
    \item For a given kernel group, find the \textit{top-m} images according to its group kernel norm value.
    \item For each kernel in the kernel group find the $IoU_c$ score for each of the \textit{top-m} images.
    \item Calculate the mean $IoU_c$ score for each kernel over all images.
    \item Calculate the mean of the mean $IoU_c$ score for each kernel to obtain the kernel Group's $IoU_c$ score. 
\end{enumerate}
Fig \ref{fig_2} illustrates the $IoU_c$ scores calculation for a single kernel.

The label of the corresponding predicate of a kernel group is chosen as the set of concepts that have their \textit{normalized} $IoU_c$ score in a certain ``margin" from the top concept. This is controlled using the user-defined \textit{margin} hyperparameter.

For example, if the $IoU_c$ score for kernel group $12$ is \{$cabinets$ : $0.5$, $door$ : $0.4$, $drawer$ : $0.1$\} then with a \textit{margin} of $0.1$ the label for the corresponding predicate will be ``$cabinets1\_door1$" since the concept \textit{door} is in the $0.1$ margin from the top concept \textit{cabinets}. Note, each concept name in the label is appended with a unique numerical identifier (in this case $1$), to distinguish it from the the other kernel groups that might learn the same concept. Say, if kernel group $25$ is also detecting $cabinets$ then its predicate's label would be ``$cabinets2\_...$" where $...$ denotes the other concepts that the kernel group $25$ might be detecting.

\section{Experiments and Results}
\noindent\textbf{Exp 1 (Setup):} We compare the performance of NeSyFOLD-G framework with that of the NeSyFOLD framework on various datasets. We report the accuracy, fidelity, number of unique predicates in the rule-set, number of rules generated and the size of the rule-set. 
Size is calculated as the total number of predicates in the bodies of the rules that constitute the logic program generated by NeSyFOLD and NeSyFOLD-G.

We used a VGG16 CNN with pre-trained weights on the Imagenet dataset \cite{deng2009imagenet}. We trained for $100$ epochs with a batch size of $32$. We used the Adam \cite{adamoptim} optimizer and applied class weights for imbalanced data. We also used $L2$ regularization of $0.005$ on all layers and a learning rate of $5 \times 10^-7$. We used a decay factor of $0.5$ and patience of $10$ epochs. Also, we resized all images to $224 \times 224$. We used $\alpha = 0.6$ and $\gamma = 0.7$ for all the datasets. For this experiment, we used the \emph{German Traffic Sign Recognition Benchmark} (GTSRB) \cite{gtsrb}, \emph{MNIST} \cite{mnist} and the Places \cite{zhou2017places} dataset.

The GTSRB dataset has $43$ classes. Each class contains multiple instances of a physical signpost and multiple images of the signpost are provided. We used a $80:20$ training-validation split per class and used the provided test set to report the performance metrics of the models.

The MNIST dataset has $10$ classes. Each class contains images of a handwritten digit from $0$ to $9$. We split the standard training set into train and validation set by using the last $10k$ images for the validation set. We used the provided test set to report the results.

The Places dataset has images of various scenes. To see the effect of varying the number of classes $\in$ \{2, 3, 5, 10\} we train on the bathroom and bedroom class (PLACES2) first. Then we add the kitchen class (PLACES3.1), then dining room, living room (PLACES5) and finally home office, office, waiting room, conference room and hotel room (PLACES10). We also selected $2$ additional subsets of $3$ classes each namely, \{desert road, forest road, street\} (PLACES3.2) and \{desert road, driveway, highway\} (PLACES3.3). We obtained the train and the test set by selecting $1k$ images from each class for the test set and the other $4k$ for the training set. We use the given validation set to tune our hyperparameters.

The NeSy-G model was created using the learning procedure described previously using the NeSyFOLD-G framework and the NeSy model was created using the NeSyFOLD framework as described in \cite{nesyfold}.
The comparison between NeSyFOLD-G and NeSyFOLD is drawn in Table \ref{tb_1}. The accuracy and fidelity are reported on the test set. The results are reported after $5$ runs on each dataset. Note, fidelity determines how closely a model follows the predictions of another model. Since the NeSy-G and NeSy models are created from the trained model they should show high fidelity w.r.t the CNN.


\begin{table*}[h!]
\centering
\setlength{\tabcolsep}{4.0pt}
\begin{tabular}{@{}rllllll@{}}
\toprule
\multicolumn{1}{l}{Data} & Algo & Fid. & Acc.& Pred. & Rules & Size \\ \midrule
\multirow{2}{*}{PLACES2}     & NF   & $\textbf{0.93} \pm \textbf{0.01}$ &$0.92\pm 0.01$ & $16 \pm 2$ & $12 \pm 2$ & $28 \pm 5$\\
                            & NF-G & $\textbf{0.93} \pm \textbf{0.0}$ &$\textbf{0.93} \pm \textbf{0.0}$ & $\textbf{8} \pm \textbf{1}$ & $\textbf{7} \pm \textbf{1}$ & $\textbf{11} \pm \textbf{2}$\\ \cmidrule(lr){1-7}                        
\multirow{2}{*}{PLACES3.1}     & NF   & $0.85 \pm 0.03$ &$0.84 \pm 0.03$ & $28 \pm 6$ & $21 \pm 4$ & $49 \pm 9$\\
                            & NF-G & $\textbf{0.87} \pm \textbf{0.01}$ &$\textbf{0.86} \pm \textbf{0.01}$ & $\textbf{20} \pm \textbf{7}$ & $\textbf{15} \pm \textbf{3}$ & $\textbf{31} \pm \textbf{9}$\\ \cmidrule(lr){1-7}                            
\multirow{2}{*}{PLACES3.2}        & NF   & $\textbf{0.94} \pm \textbf{0.0}$ &$\textbf{0.92} \pm \textbf{0.0}$ & $16 \pm 4$ & $13 \pm 3$ & $26 \pm 7$\\
                            & NF-G & $\textbf{0.94} \pm \textbf{0.01}$ &$\textbf{0.92} \pm \textbf{0.01}$ & $\textbf{12} \pm \textbf{3}$ & $\textbf{10} \pm \textbf{1}$ & $\textbf{18} \pm \textbf{3}$\\ \cmidrule(lr){1-7}
\multirow{2}{*}{PLACES3.3}        & NF   & $\textbf{0.83} \pm \textbf{0.01}$ &$0.79 \pm 0.01$ & $32 \pm 5$ & $23\pm 3$ & $60 \pm 11$ \\
                            & NF-G & $\textbf{0.83} \pm \textbf{0.01}$ &$\textbf{0.80} \pm \textbf{0.01}$ & $\textbf{30} \pm \textbf{2}$ & $\textbf{21} \pm \textbf{3}$ & $\textbf{53} \pm \textbf{6}$\\ \cmidrule(lr){1-7}

\multirow{2}{*}{PLACES5}     & NF   & $0.67 \pm 0.03$ &$0.64 \pm 0.03$ & $56 \pm 3$ & $52 \pm 4$ & $131 \pm 10$ \\
                            & NF-G & $\textbf{0.68} \pm \textbf{0.02}$ &$\textbf{0.65} \pm \textbf{0.02}$ & $\textbf{41} \pm \textbf{4}$ & $\textbf{34} \pm \textbf{6}$ & $\textbf{83} \pm \textbf{13}$\\ \cmidrule(lr){1-7}
\multirow{2}{*}{PLACES10}     & NF  & $0.23 \pm 0.19$ &$0.20 \pm 0.17$ & $\textbf{33} \pm \textbf{28}$ & $\textbf{32} \pm \textbf{27}$ & $\textbf{78} \pm \textbf{66}$ \\
                            & NF-G & $\textbf{0.33} \pm \textbf{0.17}$ &$\textbf{0.30} \pm \textbf{0.15}$ & $74 \pm 39$ & $73 \pm 39$ & $184 \pm 97$\\ \cmidrule(lr){1-7}
\multirow{2}{*}{GTSRB}     & NF   & $0.75 \pm 0.04$ &$0.75 \pm 0.04$ & $206 \pm 28$ & $134 \pm 26$ & $418 \pm 79$ \\
                            & NF-G & $\textbf{0.76} \pm \textbf{0.02}$ &$\textbf{0.76} \pm \textbf{0.02}$ & $\textbf{176} \pm \textbf{13}$ & $\textbf{98} \pm \textbf{11}$ & $\textbf{320} \pm \textbf{30}$\\ \cmidrule(lr){1-7}
\multirow{2}{*}{MNIST}     & NF   & $\textbf{0.91} \pm \textbf{0.01}$ &$\textbf{0.91} \pm \textbf{0.01}$ & $132 \pm 9$ & $90 \pm 7$ & $271 \pm 25$ \\
                            & NF-G & $0.90 \pm 0.01$ &$0.90 \pm 0.01$ & $\textbf{103} \pm \textbf{12}$ & $\textbf{79} \pm \textbf{10}$ & $\textbf{216} \pm \textbf{28}$\\ 
                            \cmidrule(lr){1-7}                                                

\end{tabular}
\caption{Comparison NeSyFOLD (NF) vs NeSyFOLD-G (NF-G).}
\label{tb_1}
\end{table*}

\medskip\noindent\textbf{Exp 1 (Result):} Table \ref{tb_1} clearly shows that the NeSy-G model outperforms NeSy model w.r.t accuracy and fidelity in most cases and is comparable otherwise. More importantly, the advantage of using the NeSyFOLD-G framework is apparent from the reduction in the number of predicates, number of rules and the overall size of the rule-set that is generated. 

The reduction in size of the rule-set is a direct indication of the improved interpretability as pointed out by Lage et al. \cite{rulesetinterpretability}. The main difference between the NeSyFOLD and the NeSyFOLD-G framework is the grouping of similar kernels in the latter. The grouped kernel forms better features in the binarization table that is generated after binarizing the group norms. The grouping helps in creating more informative features for the FOLD-SE-M algorithm to generate the rules from. Hence, in a fewer number of predicates and rules, (as compared to NeSyFOLD) the same information can be captured.

Note that as the number of classes increases as in the case of \textit{PLACES2, PLACES3.1, PLACES3.2, PLACES3.3, PLACES5} and \textit{PLACES10} both the models show a decrease in the accuracy and fidelity. This is because as the number of classes increases, more number of kernels are needed to represent the knowledge and consequently more kernels have to be binarized. Thus the loss incurred due to binarization of the kernels increases as the number of classes increases. Notice that for PLACES10 the size of the rule-set generated by NeSyFOLD-G is larger than that generated by NeSyFOLD. This is because for $2$ out of the $5$ runs, NeSyFOLD could not generate any rule-set as the FOLD-SE-M algorithm could not find good enough features in the binarization table. Due to the size of the training set being relatively large ($40k$ examples) and the large number of classes ($10$ classes), the loss due to binarization rapidly increases. This is also the reason why the accuracy and fidelity is very low.
However, since NeSyFOLD-G uses kernel grouping, the FOLD-SE-M algorithm gets to work with better features in the binarization table and thus the accuracy and fidelity is much higher compared to NeSyFOLD and thus the rule-set size is also high on average. Although in $1$ run NeSyFOLD-G also manages to find no rule-set that explains the predictions of the CNN. 

\begin{figure}[h!]
    \centering
    \includegraphics[width=1\linewidth, height = 9cm]{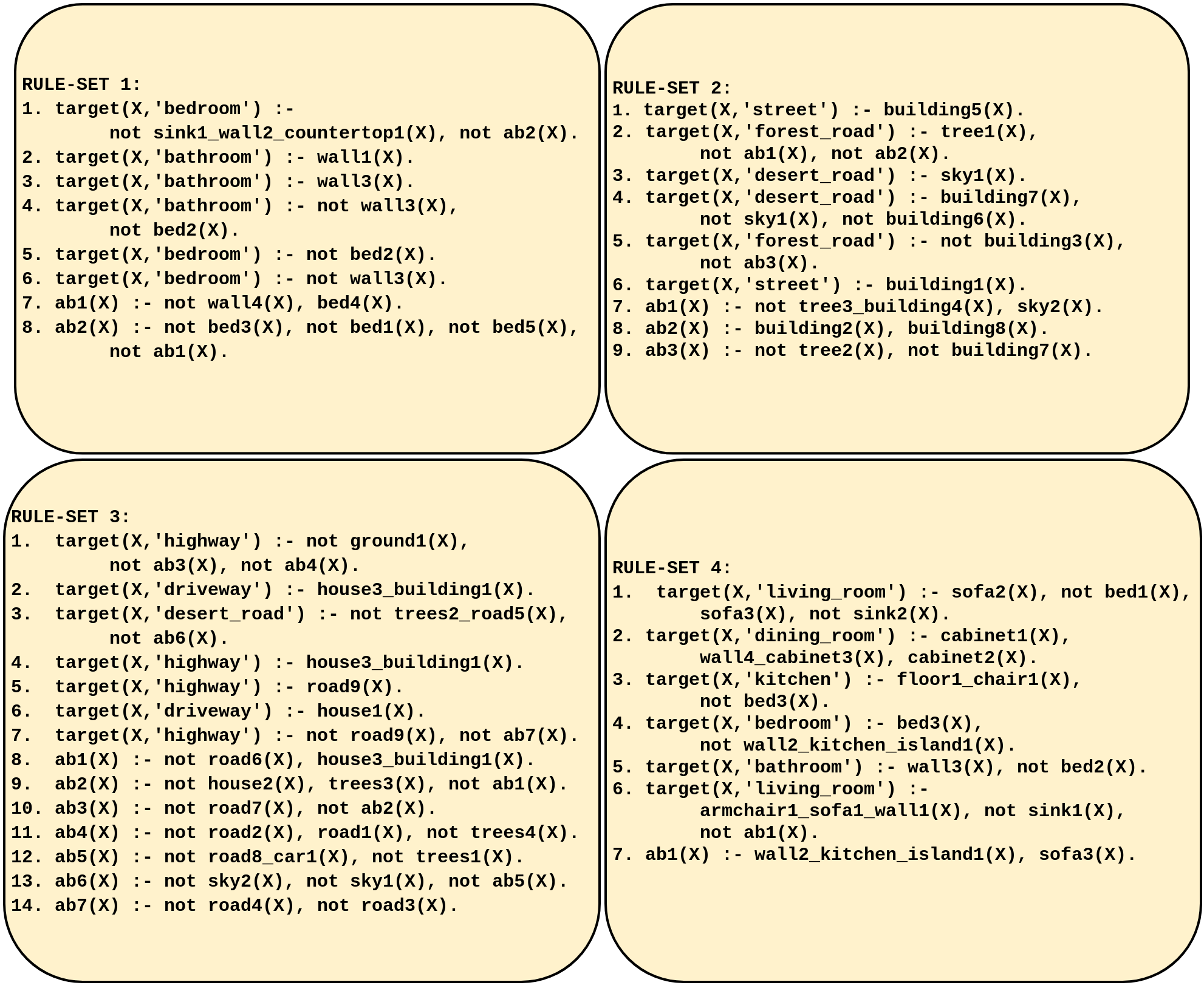}
    \caption{The labelled rule-sets generated by NeSyFOLD-G for PLACES2 (RULE-SET 1) , PLACES3.2 (RULE-SET 2), PLACES3.3 (RULE-SET 3) and PLACES5 (RULE-SET 4)}
    \label{fig_3}
\end{figure}
\begin{figure}[h!]
    \centering
    \includegraphics[width=1\linewidth, height= 5cm]{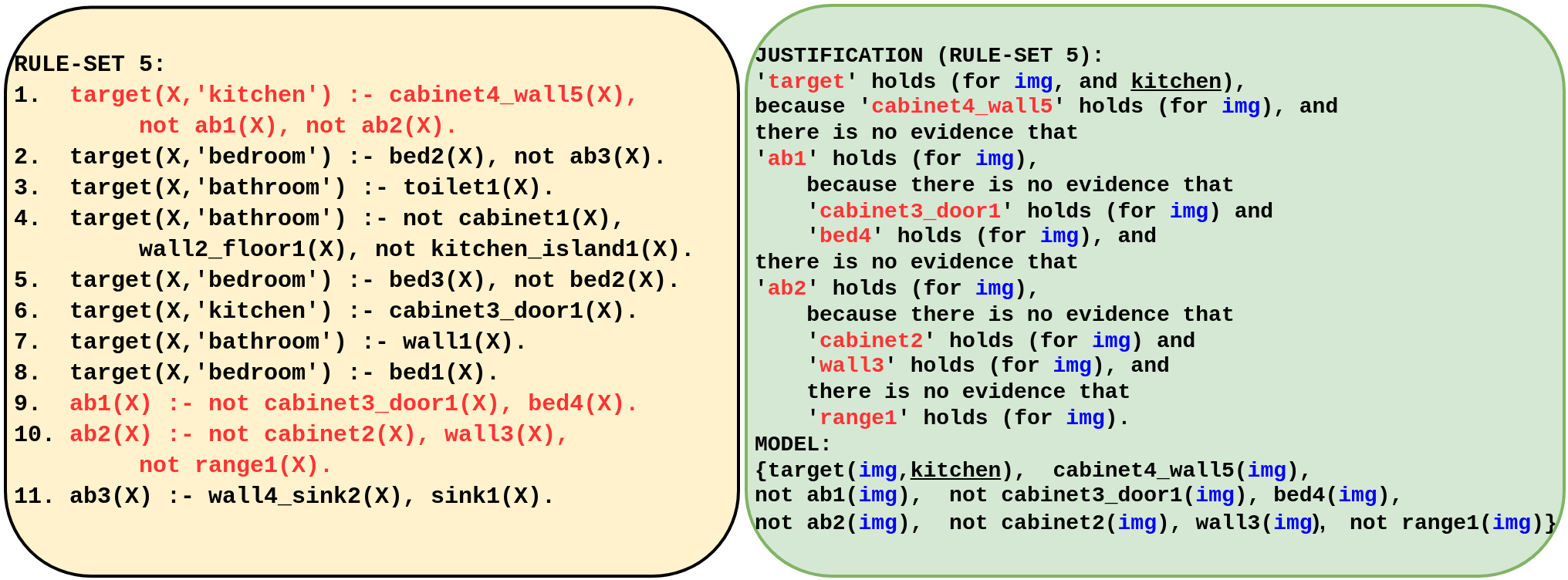}
    \caption{The justification (right) obtained from s(CASP) for an image ``img" when running the query {\tt ?- target(img, X).} against RULE-SET 5 (left).}
    \label{fig_4}
\end{figure}

\medskip\noindent\textbf{Exp 2 (setup):} We use the procedure described previously, for semantic labelling of the predicates in the rule-set generated. We use the \textit{ADE20k} dataset \cite{ade20k} in our experiments. It provides manually annotated semantic segmentation masks for a few images of all the classes of the Places dataset. The GTSRB and MNIST datasets do not have any semantic segmentation masks available. Hence, for all the subsets of classes of the Places dataset reported in Table \ref{tb_1}, we show the effect of using the semantic labelling algorithm described in Section 5. In Fig. \ref{fig_3} we have shown labelled rule-sets for the PLACES2, PLACES3.1 and PLACES3.2, PLACES3.3 and PLACES5 datasets. 
We used a $ratio$ of $0.8$ for all datasets, $tail : 5e^{-3}$ for PLACES2, PLACES3.1 and PLACES3.2, $tail: 1e^{-2}$ for PLACES3.3 and PLACES5 dataset. A similarity threshold $\theta_{s}$ of $0.8$ was used for generating the rule-sets. We used a margin of $0.05$ to label the raw rule-sets. We do not show the labelled rule-set for PLACES10 since the accuracy of the NeSy-G model is very low on the dataset. 

\medskip 
\noindent\textbf{Exp 2 (result):}
 The labelled rule-sets make intuitive sense to humans. This representation of knowledge in default theory in our opinion makes the rule-set easy to understand. The rule-set captures the knowledge of the trained CNN. For example in RULE-SET 2, the first rule states that ``an image 
{\tt X} is a `street' if there is evidence of the concept `building' in the image". Similarly the second rule states that ``an image {\tt X} is a `forest road' if there is evidence of the concept `tree' in the image and there is no evidence of some abnormal conditions `ab1' and `ab2'. 

Notice how in rules 2,3 of RULE-SET 1 in Fig. \ref{fig_3} the group of kernels, now labelled as `wall1' and `wall3' are (most probably) detecting a certain type of patterns on the walls that are indicative of bathrooms, possibly tiles. The kernels are labelled as wall only because the semantic segmentation masks available to us have the label `wall' for the pixels that denote wall in the image. Hence we are restricted to the expressiveness of the annotations available to us. This can be alleviated by labelling the predicates via manual observation.

Note that in the first rule of RULE-SET 5 (Fig. \ref{fig_4}) there is a predicate {\tt cabinet4\_wall5/1}. This predicate corresponds to the kernel group in the CNN that is detecting either both cabinets and walls separately or a specific region in the images that contains a portion of cabinets and wall. It is hard to distinguish between the two cases. 

Fig \ref{fig_4} shows a sample justification obtained from s(CASP) for some image ``img". The binarized vector associated with ``img" is used to write the facts 
 and the query {\tt target(img, X)} is executed against RULE-SET 5. The first rule (shown in red) was satisfied. The first model found by s(CASP) that satisfies the rule-set binds the value of {\tt X} to `kitchen'. Hence, the predicted class of the image ``img" is kitchen. 

\section{Related Work}
A similar approach of generating rules from the CNN was adopted by Townsend et al. \cite{eric}, \cite{townsend2022on} where they used a decision tree algorithm to generate the rule-set. However, Padalkar et al. \cite{nesyfold} have shown that using FOLD-SE-M generates a much smaller rule-set and higher accuracy and fidelity.

There is a lot of past work which focuses on visualizing the outputs of the layers of the CNN. These methods try to map the relationship between the input pixels and the output of the neurons. Zeiler et al. \cite{zeiler2014visualizing} and  Zhou et al. \cite{zhou2016learning} use the output activation while others \cite{selvaraju2017grad},\cite{https://doi.org/10.48550/arxiv.1412.6815},\cite{https://doi.org/10.48550/arxiv.1312.6034} use gradients to find the mapping. Unlike NeSyFOLD-G, these visualization methods do not generate any rule-set. Zeiler et al. \cite{zeiler2014visualizing} use similar ideas to analyze what specific kernels in the CNN are invoked. 
There are fewer existing publications on methods for modeling relations between the various important features and generating explanations from them. 
Ferreira et al. \cite{KR2022-45} use multiple mapping networks that are trained to map the activation values of the main network's output to the human-defined concepts represented in an induced logic-based theory. Their method needs multiple neural networks besides the main network that the user has to provide. 

Qi et al. \cite{Qi2021-QIEDN} propose an Explanation Neural Network (XNN) which learns an embedding in high-dimension space and maps it to a low-dimension explanation space to explain the predictions of the network. A sentence-like explanation including the features is then generated manually. No rules are generated and manual effort is needed. Chen et al.
\cite{chen2019looks} introduce a prototype layer in the network that learns to classify images in terms of various parts of the image. They assume that there is a one to one mapping between the concepts and the kernels. We do not make such an assumption. Zhang et al. \cite{zhang2017growing},\cite{zhang2018interpreting} learn disentangled concepts from the CNN and represent them in a hierarchical graph so that there is no assumption of a one to one kernel-concept mapping. However, no logical explanation is generated. Bologna et al. \cite{bologna2020two} extract propositional rules from CNNs. Their system operates at the neuron level, while NeSyFold-G works with groups of neurons. 

Our NeSyFOLD-G framework uses FOLD-SE-M to extract a logic program from the binarization table. 
There are other works that focus on extracting logic programs such as the ILASP system \cite{ilasp} by Law et al. and the XHAIL \cite{xhail} system by Ray et al. which induce an answer set program from the data however these systems do not learn rules from images. Some other works \cite{nesyilp}, \cite{diffilp}, \cite{diffilpext} use a  neurosymbolic system to induce logic rules from data. These systems belongs to the Neuro:Symbolic $\rightarrow$ Neuro category whereas ours belongs to the Neuro;Symbolic category.


\section{Conclusion and Future Work}
In this paper we have shown how the NeSyFOLD-G framework can be used to make a CNN more interpretable. We used the framework with a trained CNN to derive a NeSy-G model that constitutes the CNN with all layers after the last convolutional layer replaced by the rule-set generated by FOLD-SE-M algorithm. We compared the performance of the NeSyFOLD-G framework with that of the NeSyFOLD framework on various datasets. The major difference between the NeSyFOLD-G and the NeSyFOLD framework is that in the former, groups of similar kernels are found and the output of these groups kernels is then binarized to produce the binarization table, that is used as input to the FOLD-SE-M algorithm which generates a rule-set. The kernel grouping algorithm is a novel contribution of this work. In the NeSyFOLD framework each individual kernel's output is binarized and the rules are generated based on the binarization table thus constructed. 

We show in the experiments that grouping similar kernels leads to the creation of better features in the binarization table which consequently leads to a more succinct rule-set. The NeSyFOLD-G framework always generates a smaller rule-set than that generated by the NeSyFOLD framework while either outperforming or showing comparable accuracy and fidelity. 

We also introduced a novel semantic labelling algorithm that can be used for labelling each predicate that appears in the rule-set with the concepts(s) that its corresponding kernel group represents. We showed two labelled rule-sets and an example justification of a prediction that can be obtained using the s(CASP) ASP system.

Note that both NeSyFOLD-G and NeSyFOLD are aimed at representing the connectionist knowledge of the CNN in terms of a symbolic rule-set. The symbolic rule-set can then be scrutinized by experts to figure out the biases that the CNN might have learnt from the data and these help in avoiding spurious predictions in sensitive domains such as medical imaging. The advantage that NeSyFOLD-G provides is that the interpretability of the rule-set increases as the size of the generated rule-set is significantly smaller.

We acknowledge that the semantic segmentation masks of images may not be readily available depending on the domain, in which case the semantic labelling of the predicates has to be done manually. Our NeSyFOLD-G framework helps in this regard as well, as it decreases the number of predicates that need to be labelled.  

As the number of classes increases, the loss in accuracy also increases due to the binarization of more kernels. We plan to explore end-to-end training of the CNN with the rules generated so that this loss in binarization can be reduced during training itself.

In future, we plan to use NeSyFOLD-G for real-world tasks such as interpretable breast cancer prediction. We also intend to explore combining the knowledge of two or more CNNs by producing a single rule-set that contains the kernels of the corresponding CNNs as predicates. We also plan to investigate how the knowledge from the generated rules can be backpropagated to improve the performance of a CNN \cite{10.5555/3491440.3491683}.\\
\section*{Acknowledgments}
Authors are supported by US NSF Grants IIS 1910131 and IIP 1916206, US DoD, and various industry grants.
\bibliographystyle{splncs04}
\bibliography{main}
\end{document}